\pdfoutput=1
\documentclass[11pt]{article}
\usepackage{times}
\usepackage{latexsym}
\usepackage[T1]{fontenc}
\usepackage[utf8]{inputenc}
\usepackage{microtype}
\usepackage{inconsolata}
\usepackage{bussproofs}
\usepackage{amsmath}
\usepackage{amssymb, mathrsfs}
\usepackage{tikz}
\usepackage{pgfplots}
\usepackage{subcaption}
\usepackage{tikz-dependency}
\usepackage{hyperref}
\pgfplotsset{compat=1.17}
\usetikzlibrary{positioning}

\newcommand{\backlinks}{B_\Psi}

\newcommand{\constant}[1]{{\bf c}_{#1}}
\newcommand{\variable}[1]{{\bf x}_{#1}}

\newcommand{\xvariable}{{\bf x}}
\newcommand{\rvariable}{{\bf r}}
\newcommand{\zvariable}{{\bf z}}
\newcommand{\cvariable}{{\bf c}}
\newcommand{\avariable}{{\bf a}}
\newcommand{\yvariable}{{\bf y}}

\newcommand{\pvariable}{{\bf p}}

\newcommand{\pvariableset}{\left\{\pvariable\right\}}
\newcommand{\qvariable}{{\bf q}}
\newcommand{\gvariable}{{\bf g}}
\newcommand{\hvariable}{{\bf h}}
\newcommand{\wvariable}{{\bf w}}
\newcommand{\mvariable}{{\bf m}}
\newcommand{\condsep}{\ |\ }

\newcommand{\xjack}{\xvariable_{jack}}
\newcommand{\xjill}{\xvariable_{jill}}
\newcommand{\opand}{\textbf{\em and}}
\newcommand{\opor}{\textbf{\em or}}
\newcommand{\opxor}{\textbf{\em xor}}

\title{\bf The Quantified Boolean Bayesian Network \\ 
\vspace{10pt}
 \Large \textmd{Theory and Experiments with a Logical Graphical Model
 \thanks{The author acknowledges the use of {\em ChatGPT} in the production of this work, for research, proof-reading, and the production of equations.}}
\vspace{25pt}
}
\author{
    {\Large Greg Coppola}
    \\
    {\em coppola.ai} \\
    Research. Develop. Meme.
}
\date{February 11, 2024}
\begin{document}
\maketitle
\tableofcontents
\section{Contributions}
We introduce the {\bf Quantified Boolean Bayesian Network}, {\em QBBN} for short, a model from the {\em Bayesian Network} family \cite{pearl1988probabilistic, neapolitan2003learning}, constructed and analyzed to provide a {\em unified theory} of {\em logical} and {\em statistical} {\em reasoning}.
In particular, our work makes the following contributions:
\begin{itemize}
    \item {\bf Unified Model of Logical and Probabilistic Reasoning} \\ 
        We provide a single data structure, the {\em QBBN}, which can do both:
        \begin{itemize}
            \item {\bf Statistical Reasoning} -- The {\em QBBN} is a {\em graphical model} that can answer {\em probabilistic queries} \cite{koller2009probabilistic}, i.e. for {\em information retrieval} \cite{Shannon1948}.
            \item {\bf Logical Reasoning} -- We show how the {\em QBBN} fits precisely into a larger {\em consistent} and {\em complete} {\em logical deduction system} \cite{Gentzen1934} for the {\em first-order} calculus \cite{Frege1879}.
        \end{itemize}
        The completeness proof is outlined in \cite{Coppola2024}.
    \item {\bf A Generative Model Without Hallucinations} \\
        The {\em QBBN} shows how to create a {\em generative} model of the ({\em latent logical forms} underlying) unlabeled text.
        Like the {\em large language model} \cite{Bahdanau2014NeuralMT, Vaswani2017, radford2018improving}, the {\em QBBN} is generative, and so can be used to {\em compress} the data \cite{SutskeverObservation}.
        But, the {\em QBBN} does {\em not} {\em hallucinate}.
        It reasons consistently (i.e., ensuring that $P(x) + P(\neg x) = 1$ for all questions $x$), and can {\em explain} its reasoning in terms of {\em causality}, like any Bayesian Network can.
    \item {\bf Very Efficient Bayesian Inference} \\
        In general, inference in a Bayesian Network is intractable, i.e. $\Omega(2^N)$ for $N$ random variables \cite{neapolitan2003learning}.
        Our division of Bayesian Network nodes into \opand\ and \opor\ {\em boolean  gates}, along with our use of the unguaranteed but empirically converging {\em iterative belief propagation} \cite{murphy1999loopy, Smith2008} means that {\em inference} can now be not only tractable, but {\em very efficient}, with one full pass of approximate belief propagation requiring only time $O(N2^n)$, where $N$ is the number of network variables involved, and $n$ bounds the number of incoming connections in any \opand\ or \opor\ gate. Moreover, we discuss why it may be possible to bring the factor computation cost to $O(n)$ instead of $O(2^n)$ for each of \opand\ and \opor.
    \item {\bf Fast Versus Slow Thinking} \\
        We give, to our knowledge, the first mathematical {\em explanation} of the distinction between what has come to be known as {\em fast} versus {\em slow} thinking \cite{Kahneman2011ThinkingFast}.
        This explanation is based on {\em proof theory} of the {\em natural deduction calculus}, and accords both with our graphical formulation, as well human experience. 
        As a special case of general reasoning, we analyze {\em planning}, which task \cite{Lecun2023} has argued {\em LLM}'s do not properly support.
        While \cite{Coppola2024} contains the {\em fast versus slow} analysis, this work contains the related model details.
    \item {\bf Calculus Over Dependency Trees} \\
        Empirically, {\em labeled dependnecy trees} \cite{eisner1996three} are the easiest {\em syntactic formalism} to parse to.
        Traditionally, parsing language to a {\em complete} and {\em consistent} calculus required using the {\em first-order logic} calculus \cite{Steedman1996}, but translation to {\em literally} first-order calculus, requires an unecessary imposition of {\em positional order} on arguments that is not helpful for knowledge encoding.
        By defining a complete calculus closer to the key-value {\em labeled dependency structure}, we find it is {\em easier} to encode knowledge, and we {\em minimize} the distance between the {\em surface form} and the {\em interpretation}.
\end{itemize}
\section{Motivation}
\subsection{Large Language Models}
The {\em Quantified Boolean Bayesian Network} is introduced as a remedy the following drawbacks of the {\em large language model} \cite{Bahdanau2014NeuralMT, sutskever2014sequence, Vaswani2017,devlin2018bert, radford2018improving}.

\paragraph{Hallucinations}
While the {\em large language model} is widly popular for its ability to learn complex kinds of {\em knowledge} and even some {\em reasoning} from {\em unlabeled text},
the primary empirical user complaint with {\em large language models} is that of {\em hallucinations} \cite{SutskeverHuang2023}.
That is, a {\em large language model} can return answers that are not ``supported by the training set'' when judged by a {\em human evaluator}.
This lack of reliability greatly limits throughput, because it requires all output of a {\em large language model} to be double-checked by the user.

\paragraph{Reasoning}
Another noted problem is that the LLM does not {\em reason logically}, and does not have a {\em logically consistent world view} \cite{Steedman2022, Hinton2023}.
We propose that these two problems with {\em large language models} are directly related.
That is, the fact that a {\em large language model} will return answers unsupported by the training set is {\em because} of the fact that the {\em large language model} does not understand {\em causality}.
If a {\em knowledge data structure} {\em were} able to explain its reasoning, and only return answers based on valid reasoning, then it would be unabled to return answers unsupported by the training set, and thus unable to hallucinate.

\paragraph{Planning}
\cite{Lecun2023} has noted that one problem with {\em large language models} is that they do not seem to {\em plan} properly.
We propose to understand the ``complete'' set of deduction rules as those in \cite{Prawitz1965}, and from this perspective we can analyse planning as {\em simple inferences} (see \cite{Coppola2024} and work in preparation), along with {\em $\vee$-elimination}.
This is to say, planning is a mix of {\em forward inference}, along with {\em reasoning by cases}, and this is a precisely {\em simpler} form of reasoning than general reasoning, because it does not use all the rules of general theorem-proving.

\subsection{Cognitive Science}
We are also interested in understanding the {\em human mind} and {\em human reasoning}.
\cite{Chomsky1957SyntacticStructures} proposed to look for a {\em universal grammar} underlying all the diversity of {\em human language}.
In some sense, the {\em logical language} underlying {\em surface form}, may be the only true {\em universal language} \cite{montague_proper_treatment, Steedman1996}.
So, understanding this {\em logical language} and how it interacts with both {\em logical} and {\em probabilistic} reasoning is of central concern for those interested in {\em cognitive science}.
\section{Background}
\subsection{First-Order Logic}
\subsubsection*{Explanatory Power}
In the {\em philosophy of science} it is by now taken for granted that all of {\em mathematics} and {\em science} can be expressed in terms of {\em first-order logic} (or its {\em extensions}) (see, e.g., \cite{Pelletier2000}, and the references therein).
Thus, we say that {\em first-order logic} is {\em sufficient} to model {\em human reasoning}.
Extensions include {\em second-order logic} and {\em modal logic} \cite{Prawitz1965}, but we leave this for future work, and focus on the {\em first-order logic} for simplicity.

\subsubsection*{Language and Deduction Rules}
\paragraph{Universal Quantification and Implication}
The method of {\em universal quantification} is represented by $\forall$, and {\em implication} is represented by the $\rightarrow$ symbol.
These two work crucially together, as in expressing {\em Socrates}' classic {\em syllogism} that {\em all men are mortal}:
\begin{equation} \forall \xvariable, man(\xvariable) \rightarrow mortal(\xvariable) \end{equation}
This single rule licenses an {\em unbounded} number of inferences.
For example, $man(\cvariable_{jack})$, we can conclude $mortal(\cvariable_{jack})$, and if $man(\cvariable_{arjun})$ we can conclude $mortal(\cvariable_{arjun})$, etc.
This is how in language we make {\em infinite use of finite means}, as discussed in Section \ref{s:predgraph:infinite}.

\paragraph{Logical Connectives}
There are two logical connectives designated as {\em boolean} in our system, corresponding to the two operations generally in a {\em boolean algebra}.
\paragraph{\opand}
The first connective is {\em and}, represented with $\wedge$, as in:
\begin{equation} man(\constant{jack}) \wedge mortal(\constant{jack}) \end{equation}
This means that {\em both} $man(\constant{jack})$ {\em and} $mortal(\constant{jack})$ are true.
\paragraph{\opor}
The second connective is {\em or}, represented with $\vee$, as in:
\begin{equation} man(\constant{jack}) \wedge mortal(\constant{jack}) \end{equation}
This means that {\em at least one of} the {\em terms} is true, maybe {\em both}.

\paragraph{Negation}
The negation of a statement is represented in most logical presentations using the symbol $\neg$.
For example to say {\em God is not mortal} we can write $\neg mortal(\cvariable_{God})$.
In our network, the concept of {\em negation} plays a crucial role, but there is no specific {\em junction} for negation, because each {\em boolean} variable represents a probability both for {\em true} and {\em false}.

\subsubsection*{Completeness and Consistency}
For any logical calculus, we have a notion of what is {\em provable} in that calculus.
This is evaluated against a {\em model interpretation}, that says what is {\em true}.
A logic is {\em consistent} if whatever is {\em provable} is {\em true}.
A logic is {\em completeness} if whatever is {\em true} is {\em provable}.
\cite{Godel1930} proved the consistency and completeness of first-order calculus.
A fundamental insight of this work is that, we are free to work in a more practical formalism, i.e. a {\em graphical statistical model} over {\em semantic roles}, than the {\em first-order logic} if we will only prove the {\em consistency} and {\em completeness} of this new logic, which we outline in \cite{Coppola2024} and work in preparation.

\subsection{Bayesian Networks}
\subsubsection*{Markov Graphical Models}
A distribution \( P([\pvariable_1, ..., \pvariable_N]) \) \textit{factorizes according to a factor graph \( G_F \)} if there exists a set of {\em factors} $\left\{\alpha\right\}_F$ and {\em factor functions} \( \Psi_\alpha \) such that \( P([\pvariable_1, ..., \pvariable_N]) \) can be written as \cite{Sutton2011}:
\begin{equation}
    P([\pvariable_1, ..., \pvariable_N]) = Z^{-1} \prod_{\alpha \in F} \Psi_\alpha(\left\{\pvariable\right\}_\alpha)
\end{equation}
Here, $\left\{\pvariable\right\}_\alpha$ are the set of all {\em variables} $\pvariable$ in the factor $\alpha$ and \( Z \) is a {\em normalization} constant that ensures that the probabilities sum to one.
Doing {\em normalization}, and relatedly {\em marginalization}, in a general graphical model takes time $\Omega(2^N)$.

\subsubsection*{Boolean Network}
The {\em QBBN} is deliberately formulated as a {\em boolean} network, in which all propositional variables $\pvariable$ are modeled as either taking the value {\em true}, represented by $1$, or {\em false}, represented by $0$.
Note that, while $P(\pvariable = z)$ is a probability, for $z \in \left\{0, 1\right\}$, the possible {\em values} that $\pvariable$ can take are boolean.

\subsubsection*{Markov Logic Network}
\cite{richardson2006markov} use a graphical boolean statistical network to score sentences constrained by the deductions of the first-order calculus.
Inference in {\em Markov Networks} in general is {\em \#P-complete} \cite{Roth1996HardnessApproxReasoning}, which is $\Omega(2^N)$.
Because exact inference is intractable, \cite{richardson2006markov} use approximate inference via {\em Markov Chain Monte Carlo} \cite{gilks1996markov} sampling.
We propose a model to a similar effect, but with a large improvement in run-time through the use of unguaranteed {\em belief propagation}.
This is useful because of the number of daily inferences that are currently demanded of {\em large language models}.

\subsubsection*{Traditional Bayesian Networks}
\paragraph{Directed Acyclic Graph}
A traditional {\em Bayesian Network} is a {\em directed} graphical model, where each factor maps $\alpha$ {\em input variables} $\avariable_i$ to an {\em output variable} $\zvariable$:
\begin{equation} \Psi(\zvariable \condsep \avariable_1, ..., \avariable_n) \end{equation}
For each pair $(\zvariable, \avariable)$, we will refer to $\zvariable$ as the {\em child} (or {\em conclusion}), and to $\avariable$ a the {\em parent} (or {\em assumption}).
The directed nature of the factor gives rise to two clear inference directions: {\em forwards}, in which information passes from {\em causes} to {\em effects}, and {\em backwards}, in which information passes from {\em effects} (the {\em observations}), backwards to {\em causes} (a {\em hypothesis}).

\paragraph{Complexity}
Inference in general Bayesian Networks is also {\em \#P-complete} \cite{Cooper1990}, and even {\em NP-hard} to {\em provably approximate} \cite{Roth1996HardnessApproxReasoning}.
The difficulty is owing to the difficulty of marginalizing over {\em undirected cycles} in the factor graph \cite{neapolitan2003learning,koller2009probabilistic}.
Nevertheless, {\em loopy belief propagation}, which we will henceforth call {\em iterative belief propagation}, while {\em not} provably convergent, has been found to empirically to converge in many situations \cite{murphy1999loopy, Smith2008}.
The complexity of belief propagation is discussed in detail in Section \ref{sec:complexity}.

\subsubsection*{Quantification in Bayesian Networks}
An analog of {\em universal quantification} has been studied under the rubric of {\em plate models} \cite{koller2009probabilistic}, in which nodes sharing a {\em template} structure can share weights.
We also employ this {\em parameter sharing}, but view it instead from a logical perspective as {\em quantification}.
\section{A Novel Calculus Over Semantic Roles}
\subsection{Motivation}
We have said that the calculus of {\em first-order logic} is {\em complete}, {\em consistent}, and {\em sufficient} for expressing {\em mathematics} and {\em science}.
However, the {\em language} of the {\em first-order logic} logic is quite far from the {\em labeled dependency parses} that are most easily parsed to \cite{eisner1996three, mcdonald2005non, zhang2011transition}.
For example, \cite{Lewis2013} shows how the sentence {\em Shakespeare wrote Macbeth} can be translated via a system of {\em functional categories} to a {\em first-order language} formula:
\begin{equation}
    wrote_{arg_0:\textsc{per}, arg_1:\textsc{book}}(\cvariable_{Shakespare}, \cvariable_{Macbeth})
\end{equation}
Our observation is that the written order here is an {\em artifact} of the fact, traditionally, first-order logic was done by {\em writing on a page}, and that for computational purposes, the {\em written order} of the arguments is irrelevant, given their {\em argument labels}.
Instead, we use a {\em key-valued calculus} formalism like:
\begin{equation}
(wrote, \left\{ arg_0: \cvariable_{Shakespare}, arg_1:\cvariable_{Macbeth}\right\})
\end{equation}
That is, it is easier to {\em ignore the order} of the arguments, and use a {\em key-value} map to index the arguments.
In practice, as we will see, it is easier to encode implications if we ignore the order, and only use the {\em function name} and the {\em labeled key-value} pairs.
Also, this formulation matches the way that an {\em attention node} works, in that an attention function can be described as mapping a query and a set of key-value pairs to an output \cite{Vaswani2017}.
In the attention network, these objects are all {\em vectors}, while here they are {\em symbols}.
To be clear, we are not claiming there is {\em no} book-keeping to do to get from {\em surface structure} to {\em logical} structure.
We can associate each of the {\em labeled dependencies} each with a {\em function application} fom {\em categorial grammar} \cite{BarHillel1953}.
However, the pipeline can be greatly simplified on the {\em parsing side} and also on the {\em knowledge representation} side if we {\em feel free} to invent more {\em flexible} logical calculi (e.g., {\em key-valued}), so long as we prove {\em consistency}, {\em completeness} and {\em sufficiency}.

\subsection{Language Definition}
\begin{figure}[h!]
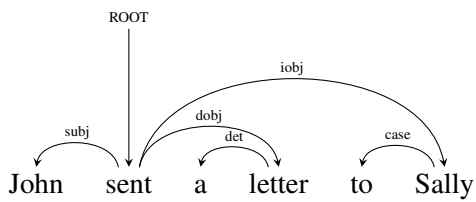

    \centering
\begin{dependency}[theme = simple]
    \begin{deptext}[column sep=1em]
       John \& sent \& a \& letter \& to \& Sally \\
    \end{deptext}
    \deproot{2}{ROOT}
    \depedge{2}{1}{subj}
    \depedge{4}{3}{det}
    \depedge{2}{4}{dobj}
    \depedge{6}{5}{case}
    \depedge{2}{6}{iobj}
\end{dependency}
\caption{A labeled dependency parse. Without labels, we could not do semantics, so this is the most simple structure that can support semantics.}
\label{fig:dependency}
\end{figure}

\paragraph{A Key-Value Calculus}
Assume we have access to a {\em labeled dependency parse} as in Figure \ref{fig:dependency}.
From this parse we can through some {\em syntactic analysis} extract a {\em proposition} of the {\em rough form}:
\begin{equation}
(\textsc{send}, \left\{
\begin{aligned}
&\textsc{subj}: \text{John}, \\
&\textsc{dobj}: \text{a letter}, \\
&\textsc{iobj}: \text{Sally}
\end{aligned}
\right\})
\label{eq:dep-semantics}
\end{equation}
By defining a predicate as close to the bare dependency structure as possible, we obviate the need to manage the book-keeping to enforce an {\em arbitrary linear order} on the arguments as in:
\begin{equation} 
    send_{subj,dobj,iobj}(\text{John}, \text{a letter}, \text{Sally})
\end{equation}
We still use a first-order style sometimes in the text to save space where the intended key-value translation is hopefully clear.

\paragraph{Truth Values}
There are two {\em boolean truth values}, {\em true}, which we write as $1$ and {\em false}, which we write as $0$.
The {\em nodes} of primary interest in {\em queries} to our {\em graphical model} are about the {\em values} of {\em propositions}, usually denoted $\pvariable$.
We can {\em query} the probabilities $P(\pvariable = 1)$ and $P(\pvariable = 0)$.
That is, we assume that each {\em proposition} is either definitely {\em true} or {\em false}, and we do not know which, but we can assign a probability in $[0, 1]$.

\paragraph{Entities}
An {\em entity} is identified by a {\em string} $e$ and corresponds to an object in our {\em information retrieval} database, e.g., {\em Taylor Swift}, {\em Beyonce}, {\em USA}, {\em China}.

\paragraph{Types} A {\em type} $\tau$ is identified by a {\em string}.
We will assume that each {\em entity} {\em exhibits} a non-negative number of {\em types}.
In {\em information retrieval} some relevant types are {\em business}, {\em individual}, {\em group}, {\em book}, or {\em product}.
Usually the type is clear from context and we will usually not write $\tau$.

\paragraph{Constants}
A {\em constant} (or {\em constant reference}) is a pair $(e, \tau)$ of {\em entity identifier} and {\em type}.
The constant refers to a specific entity.
For example, the entity \textsc{usa} exhibits the type \textsc{country}, so its constant reference would be:
\begin{equation} \constant{usa} = \textbf{constant}(\textsc{usa}, \textsc{country})\end{equation}

\paragraph{Variables}
A {\em variable} is defined by a type $\tau$.
\begin{equation} \variable{country} = \textbf{variable}(\textsc{country})\end{equation}
A {\em variable} can be {\em instantiated} by any {\em constant} of the same type.

\paragraph{Function Names}
A {\em function name} $f$ is a {\em string}, e.g., $\textsc{like}$ or $\textsc{date}$.

\paragraph{Arguments}
An {\em argument} $a_\tau$ is an object that wraps {\em either} a {\em constant} $\cvariable_\tau$ or a {\em variable} $\xvariable_\tau$.
Given an argument, we can tell which type of object it wraps ($\cvariable_\tau$ or $\xvariable_\tau$), and also recover the wrapped object.

\paragraph{Role Labels}
Each {\em role label} $r$ is a {\em string} from a {\em bounded set}, e.g. \textsc{subj}, \textsc{dobj} or \textsc{iobj}.
The role label indexes the {\em argument position} that an {\em argument} plays for a {\em function}.
A {\em labeled argument} is a pair $(r, a)$ of {\em role} and {\em argument}.

\paragraph{Role Sets and Maps}
A set of {\em roles} $\rvariable = \left\{r\right\}_{r \in \rvariable}$ is called a {\em role set}.
A {\em role-argument mapping} is a map $\mvariable_\rvariable = \left\{(r, a)\right\}_{r \in \rvariable}$ with {\em role set} $\rvariable$.
The {\em open roles} in $\mvariable$ are those pair $(r, a)$ where $a$ wraps a {\em variable}.
The {\em filled roles} are those where $a$ wraps a {\em constant}.

\paragraph{Predicates}
A {\em predicate}'s {\em type} is {\em defined} by pair of a {\em function name} and a set of {\em roles labels}.
\begin{equation} \tau(\qvariable) = (f, \rvariable)\end{equation}
A {\em predicate instance} is a pair of a {\em function name} and a {\em role-argument mapping}:
\begin{equation} \qvariable = (f, \mvariable_\rvariable)\end{equation}
An example of a predicate is:
\begin{equation} \qvariable = (\textsc{like}, \left\{\textsc{sub}: \variable{jack}, \textsc{obj}: \variable{jill} \right\})\end{equation}
$\qvariable$ does {\em not} have a truth value, and we {\em cannot} ask $P(\qvariable = 1)$, because of the presence of {\em open roles} and so {\em unbound variables} $\variable{jack}$ and $\variable{jill}$.
Only when these variables are replaced by {\em constants}, referring to {\em specific entities}, will we have a truth value to estimate a probability for.

\paragraph{Propositions}
A {\em predicate} with {\em zero} open roles is called a {\em proposition}, usually denoted $\pvariable$, e.g.
\begin{equation} \pvariable = (\textsc{like}, \left\{\textsc{sub}: \constant{jack1}, \textsc{obj}: \constant{jill1} \right\})\end{equation}
Having no {\em open roles}, a {\em proposition} is {\em fully grounded} and so has a {\em probability}, and we can ask $P(\pvariable = 1)$.
E.g., in this case, we can ask whether $like(\constant{jack1}, \constant{jill1})$ in particular.

\subsection{Quantification and Implication}
\subsubsection{Statistical Inference}
In the traditional first-order calculus $\forall x, A(x)\rightarrow B(x)$ means that $B$ {\em always} follows $A$.
We want to generalize $\forall$ with a {\em statistical} notion $\Psi x, A(x)\rightarrow B(x)$, which means, {\em more generally}, that $B$ follows $A$ {\em with some probability}.
Then, we have the option to {\em estimate} $\Psi$ from {\em data}.
\subsubsection{Predicate Implication Links}
\label{sec:running_example}
\paragraph{Example}
We will introduce the running example of {\em binary dating}, in which we have a {\em bipartite graph} with {\em two} types of entities, those of type $\xjack$ and those of type $\xjill$, and we have a predicate of interest:
\begin{equation} (\textsc{date}, \left\{ \textsc{subj}: \xjack, \textsc{dobj}:\xjill\right\}) \end{equation}
This returns true if $\xjack$ is dating $\xjill$.
Now if $\xjack$ {\em likes} $\xjill$, they are more likely to {\em date}.
We can represent this in our key-value calculus as:
\begin{equation}
\Psi[\xjack, \xjill]\left(\textsc{like}, \left\{
    \begin{aligned}
    &\textsc{subj}: \xjack, \\
    &\textsc{dobj}: \xjill, \\
    \end{aligned}
    \right\} \right)
    \rightarrow \left(
    \textsc{date}
\left\{
    \begin{aligned}
    &\textsc{subj}: \xjack, \\
    &\textsc{dobj}: \xjill, \\
    \end{aligned}
    \right\}
    \right)
\label{eq:like-date-same}
\end{equation}
We can also represent the related link that $\xjack$ and $\xjill$ are more likely to date if $\xjill$ {\em likes} $\xjack$:
\begin{equation}
\Psi[\xjack, \xjill]\left(\textsc{like}, \left\{
    \begin{aligned}
    &\textsc{subj}: \xjill, \\
    &\textsc{dobj}: \xjack, \\
    \end{aligned}
    \right\} \right)
    \rightarrow \left(
    \textsc{date}
\left\{
    \begin{aligned}
    &\textsc{subj}: \xjack, \\
    &\textsc{dobj}: \xjill, \\
    \end{aligned}
    \right\}
    \right)
\label{eq:like-date-change}
\end{equation}

\paragraph{Role Set Mapping}
Comparing \ref{eq:like-date-same} to \ref{eq:like-date-change}, we see that \ref{eq:like-date-same} maintains the same role-argument assignments in premise as conclusion:
\begin{equation}
    \left\{\textsc{subj} : \textsc{subj}, \textsc{dobj} : \textsc{dobj} \right\}
\end{equation}
In \ref{eq:like-date-change}, the roles are reversed:
\begin{equation}
    \left\{\textsc{subj} : \textsc{dobj}, \textsc{dobj} : \textsc{subj} \right\}
\end{equation}
In order to allow both possibilities, between any conclusion $\qvariable_c$ and premise $\qvariable_a$, we introduce the {\em role set mapping}, which is a map $\left\{r,  s\right\}$, where each entry $(r, s)$ indicates that the argument for role $r$ in $\qvariable_a$ should be used to fill role $s$ in $\qvariable_c$.

\paragraph{Predicate Implication Link}
A single {\em predicate implication link} is a triple:
\begin{equation}
    \Psi(\qvariable_a, \qvariable_c, \left\{r, s\right\})
\end{equation}
where $\qvariable_a$ and $\qvariable_c$ are predicates, and where $\left\{r, s\right\}$ is an appropriate role mapping between the two.
In our current implementation, we require that all open roles in each of $\qvariable_a$ and $\qvariable_c$ be filled, and that $\qvariable_a$ have less than or equal to the number of open roles of $\qvariable_c$.

\subsubsection{Conjoined Predicate Implication}
\paragraph{Motivation}
At a high level, the {\em implication links} correspond to patterns of {\em features} that we can {\em train} and {\em reuse} over {\em proposition factors}.
Suppose we want to use a {\em linear} model for these features, either because it is interpretable or because it is faster.
The problem with linear models, in general, is that they cannot separate all functions.
For example, \opxor\ cannot be separated, if the problem is interpreted naively \cite{minsky1969perceptrons}. 
However, e.g., \opxor\ {\em can} be separated if we are allowed to {\em conjoin} (or {\em combine}, or {\em take a boolean combination of}) the input features.
In the case of {\em dating}, $\xjack$ and $\xjill$ will in a modern context only date if they {\em both} like {\em each other}.
To represent this, we want a feature that only fires if {\em both} $like(\xjack, \xjill)$ {\em and} $like(\xjill, \xjack)$, i.e.:
\begin{equation}
    \Psi[\xjack, \xjill]\left[ \left\{like(\xjack, \xjill) \wedge like(\xjill, \xjack)\right\} \rightarrow date(\xjack, \xjill)\right]
    \label{eq:jack_and_jill}
\end{equation}
We discuss the role of {\em conjunction} in producing {\em higher-level features} in Section \ref{seq:conjunction-nodes}.

\paragraph{Formulation}
Where $\qvariable_{a_i}$ are each {\em predicates}, we use $\hvariable$ as short-hand for a {\em group} ({\em ordered list}) of predicates:
\[ \hvariable_a = [\qvariable_{a_1}, ..., \qvariable_{a_n}]\]
Where $\qvariable_c$ is a {\em conclusion predicate} define the {\em conjoined implication} $\Psi(\hvariable_a, \qvariable_c)$ as:
\begin{equation}
    \Psi(\hvariable_a, \qvariable_c) = \left[ (\hvariable_{a_1}, \qvariable_c, \left\{r, s\right\}_{a_1}) \wedge ... \wedge (\hvariable_n, \qvariable_c, \left\{r, s\right\}_{a_n})\right]
    \label{eq:conjoined_predicate}
\end{equation}
Here, we assume that each $\left\{r, s\right\}_{a_i}$ is appropriate to match the open roles of $\hvariable_{a_i}$ to $\qvariable_c$.
The form \ref{eq:conjoined_predicate} allows us to state an inferential link like \ref{eq:jack_and_jill}.

\section{The Proposition Graph}
\subsection{Markov Assumption}
The essential feature of a graphical model is that it makes a {\em Markov assumption}, in which each variable in the graph is independent of all nodes, given the values of its {\em neighbors}.
Because the edges are {\em directed}, the {\em neighbors} of a node are its {\em parents} and its {\em childen}.
In our Bayesian Network, each factor in the graph has the form:
\begin{equation} \Psi(\zvariable \condsep \zvariable_{a_1}, ..., \zvariable_{a_n}) \end{equation}
In this case, we would say that $\zvariable$ is the {\em child} of each $\zvariable_{a_i}$ and each $\zvariable_{a_i}$ is a {\em parent} of $\zvariable$.
Conversely, $\zvariable$ can also have effects on {\em its} children as in:
\begin{equation} \Psi(\zvariable_c \condsep \zvariable, \zvariable_{b_1}, ..., \zvariable_{b_{n-1}}) \end{equation}
Here, the $\zvariable_{b_i}$ are other parents of $\zvariable_c$.
The {\em Markov} assumption says that we can know everything we need to know about $\zvariable$ if we know the values of its {\em parents} and its {\em children}, i.e., $\zvariable$ is {\em independent} of all other nodes in the network, given its neighbors.

\subsection{Lazy Graph Storage}
\paragraph{The Full Graph is Unbounded}
For an unbounded set of {\em entities}, there are an unbounded number of {\em possible propositions} $\pvariable$, many of which will never be relevant.
For example, consider the predicate of {\em is President of the United States}.
This only applies in practice to one person, but could, in principle, apply to billions.
Thus, storing all {\em possible} propositions in hard disk memory would not be possible.
\paragraph{Stored vs. Dynamically Calculated Probabilities}
We have two options in the system for estimating the probability of a proposition $\pvariable$.
The first is that the probability for the proposition is {\em alredy computed} before the {\em query} is issued.
In the example of {\em $\xvariable_{person}$ is the President of the United States}, this can be set to {\em true} in long-term storage for the unique individual who occupies this slot.
For anyone {\em else}, we can use {\em generic} reasoning, like {\em there is only one President}, and {\em right now the President is someone else}, etc., to answer {\em no}.
\paragraph{Use of the Markov Assumption}
During training of $\Psi_\opor$, we only train {\em local factors} assuming {\em fully observed data}, which does not require a full proposition graph to be created, but only the relevant factors (see Section \ref{s:predgraph}) to be identified.
During inference, we create the proposition graph {\em dynamically} at run time from the {\em implication graph}, described in Section \ref{s:predgraph}.
Because of the {\em Markov Assumption}, and the mechanics of {\em universal quantification}, for any proposition $\pvariable$, we can determine exactly which other propositions are relevant to determining $\pvariable$.

\subsection{Boolean Algebra}
\begin{figure}[t]
    \centering
\begin{tikzpicture}
    \def\levelOne{2}
    \def\levelTwo{0.5}
    \def\levelThree{-1}
    \def\levelFour{-2.5}
    \def\levelFive{-4}
    \node (lonely) at (0,\levelOne) {$\Psi_\opor$[boy lonely]};
    \node (exciting) at (3,\levelOne) {$\Psi_\opor$[girl exciting]};
    \node (interLonely) at (0,\levelTwo) {$\Psi_\opand$};
    \node (interExciting) at (3,\levelTwo) {$\Psi_\opand$};
    \node (boyLikesGirl) at (1.5,\levelThree) {$\Psi_\opor$[boy likes girl]};
    \node (girlLikesBoy) at (4.5,\levelThree) {$\Psi_\opor$[girl likes boy]};
    \node (conjunction) at (3,\levelFour) {$\Psi_\opand$};
    \node (dates) at (3,\levelFive) {$\Psi_\opor$[boy dates girl]};
    \draw[->] (lonely) -- (interLonely);
    \draw[->] (exciting) -- (interExciting);
    \draw[->] (interLonely) -- (boyLikesGirl);
    \draw[->] (interExciting) -- (boyLikesGirl);
    \draw[->] (boyLikesGirl) -- (conjunction);
    \draw[->] (girlLikesBoy) -- (conjunction);
    \draw[->] (conjunction) -- (dates);
\end{tikzpicture}
  \caption{A {\em boolean network} that {\em alternates} between \opand\ and \opor\ gates.}
  \label{fig:alternating_network}
\end{figure}
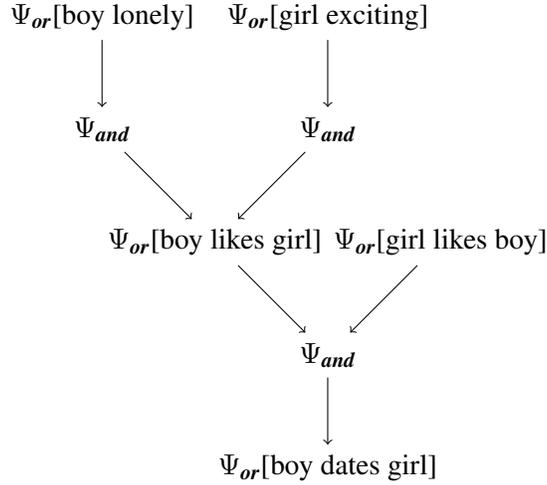

For reasons of logical completeness, and also computational efficiency, we split the graph into two kinds of {\em junctions}, or {\em factor types}:
\begin{enumerate}
    \item {\em conjunction} factors, denoted $\Psi_\opand$
    \item {\em disjunction factors}, denoted $\Psi_\opor$.
\end{enumerate}
The computation in the graph alternates between these: a {\em conjunction} factor $\Psi_\opand$ feeds into a {\em disjunction} factor $\Psi_\opor$, and vice versa, as depicted in Figure \ref{fig:alternating_network}.

\subsection{Bipartite Graph}
Because the factor types $\Psi_\opand$ and $\Psi_\opor$ always alternate, we have a {\em bipartite graph}.
Suppose $\pvariable_1, ..., \pvariable_n$ are each {\em propositions}.
Then we say 
\begin{equation} \gvariable = \left\{\pvariable_1 \wedge ... \wedge \pvariable_n\right\} \end{equation}
is a {\em proposition group}, which are interpreted as {\em conjoined}.
Then, the two types of variables in the graph then are:
\begin{enumerate}
    \item $\pvariable$, which represents a {\em single proposition}
    \item $\gvariable$, which represents a {\em conjoined proposition group}
\end{enumerate}
For many purposes in the graphical model (e.g., {\em message passing} calculuations), we can abstract over whether a node is $\gvariable$ and $\pvariable$, and we refer to {\em generic graphical nodes} as $\zvariable$.
It is to be understood that each $\zvariable$ actually {\em wraps} a $\gvariable$ or a $\pvariable$, and that we can get either the underlying type or underlying value from any $\zvariable$ at any time.

\subsection{Conjunction Nodes}
\label{seq:conjunction-nodes}
\subsubsection{Deterministic Definition}
The {\em conjunctive} factor $\Psi_\opand$ is defined in terms of the {\em \opand\ gate}:
\begin{equation} \opand(\pvariable_1, ..., \pvariable_n) = \pvariable_1 \wedge ... \wedge \pvariable_n \end{equation}
Then:
\begin{equation}
    \Psi_{\opand}(\gvariable \condsep \pvariable_1, \ldots, \pvariable_n) = 
    \begin{cases} 
        1 & \text{if } \gvariable == \opand(\pvariable_1, \ldots, \pvariable_n), \\
        0 & \text{otherwise}
    \end{cases}
    \label{eq:op_and_determinstic}
\end{equation}
This is used in the completeness proof, but also in learned models.

\subsubsection{Higher-Level Features}
Let us meditate on the fact that the $\Psi_\opand$ factor is {\em always} deterministic, i.e., we do not train this even when we are interested in statistical inference.
One way to justify this is that, intuitively, \opand's role is to create {\em higher-level features}, between which we can learn relationships.
This is like a {\em discrete} analog to the {\em higher-level} features that {\em multi-layer networks} learn \cite{Rumelhart1986, LeCun1989}.
For example, suppose we are given the information about $like(\xjack, \xjill)$ and $like(\xjill, \xjack)$ as {\em simple} features to predict $date(\xjack, \xjill)$.
Supposing the relevant higher level feature is $like(\xjack, \xjill) \land like(\xjill, \xjack)$, one option is run these features through a multi-layer network, which will be able to {\em learn} this feature, in a {\em differentiable} way.
But, this higher-level feature emerges as an effectively {\em emergent} behavior.
This leads to the problem of {\em interpretability} \cite{Hinton2015, Ribeiro2016, Lundberg2017}.
The {\em QBBN} is another interpretation of {\em interpretability}, because the features must be {\em explicitly} conjoined in order to work.
The problem is not one of {\em interpreting} the model, but {\em constructing} the model in the first place, since the individual {\em function names} and {\em role labels} underlying logical ``language'' are latent \cite{Steedman1996}, and presumably would be discovered through something analogous to {\em category splitting} in a generative model \cite{Petrov2006}.

\subsection{Disjunction Nodes}
\subsubsection{Deterministic Definition}
The deterministic {\em disjunctive} factor $\Psi_{\opor}$ used for the {\em completeness proof} (also see \cite{Coppola2024}), is defined in terms of the {\em \opor\ gate}:
\begin{equation} 
    \opor(\gvariable_1, \ldots, \gvariable_n) = \gvariable_1 \vee \ldots \vee \gvariable_n 
\end{equation}
The {\em deterministic} version of \opor, used in the {\em completeness} proof, and can be used any time we want exact logical \opor, is defined as:
\begin{equation}
    \Psi_{\opor}(\pvariable \condsep \gvariable_1, \ldots, \gvariable_n) = 
    \begin{cases} 
        1 & \text{if } \pvariable == \opor(\gvariable_1, \ldots, \gvariable_n), \\
        0 & \text{otherwise}
    \end{cases}
    \label{eq:op_or_determinstic}
\end{equation}
When interested in {\em statistical inference}, we {\em learn} this model, as discussed in Section \ref{sec:learned_model}.

\subsubsection{Learned Disjunctive Model}
\label{sec:learned_model}

For the {\em learned model}, we model $\Psi_\opor$ using {\em linear exponential} model.
For a boolean variable \( \pvariable \) with boolean features \( \gvariable_1, ..., \gvariable_n \), the {\em factor potential} has the form:
\begin{equation} 
\Psi_{\opor} (\pvariable \condsep \gvariable_1, ..., \gvariable_n) = \exp{ \left\{\sum_{i=1}^{n} {\wvariable \cdot \phi(\pvariable, \gvariable_i)}\right\}}
\label{e_linear_exponential}
\end{equation}
Here, \( \wvariable \) is a weight vector, and \( \phi(\pvariable, \gvariable_i) \) is a feature discussed in Section \ref{s:feature_function}.
The probability \( P(\pvariable \condsep \gvariable_1, ..., \gvariable_n) \) is obtained by normalization over the two possible values for \( \pvariable \in \left\{0, 1\right\} \):
\begin{equation} 
P(\pvariable = p \condsep \gvariable_1, ..., \gvariable_n) = \frac{\Psi_{\opor} (p \condsep \gvariable_1, ..., \gvariable_n)}{\Psi_{\opor} (1 \condsep \gvariable_1, ..., \gvariable_n) + \Psi_{\opor} (0 \condsep \gvariable_1, ..., \gvariable_n)} 
\end{equation}

\subsubsection{The Similarity Between Disjunction and Linear Exponential}
To underline the similarity between the {\em linear exponential model} and {\em disjunction} $\Psi_\opor$, consider how we implement $\opor$ using a log-linear model.
That is, the dependence of $Y$ on $X_1$ and $X_2$ can be expressed as:
\begin{equation}
    P(Y=1|X_1, X_2) = \frac{1}{1 + \exp(-(\beta_0 + \beta_1X_1 + \beta_2X_2))}
\end{equation}
If we set $\beta_0 = -0.5$, a negative bias, and $\beta_1 = \beta_2 = 1$, then this predicts $Y = 1$ if either $X_1 = 1$ or $X_2 = 1$, but $Y=0$ otherwise.
That is, it implements \opor, and this technique scales for $n>2$.

\subsubsection{On the Use of a Linear Model}
One might ask whether it is {\em simplistic} to use a {\em linear} model for any reason when we have availble advanced networks like {\em multi-layer networks} and {\em attention}, etc.
The use of non-linear networks in this context can be investigated.
However, we reiterate it is the role of the {\em conjunction} gates to create the {\em higher-level} features that are accomplished currently with {\em multi-layer networks} \cite{Rumelhart1986}.
Linear weights are easily {\em interpretable}, which is good for {\em human-computer alignment}.
Also, for certain definitions of $\Psi_\opor$, like the {\em Noisy Or} gate discussed in Section \ref{sec:noisy_or}, updates can be {\em fast}, i.e. $O(n)$ instead of $O(2^n)$, because of the independence of inputs.
We leave it to future work to decide whether any $O(n)$ models for $\Psi_\opor$ are useful in practice.

\section{The Implication Graph}
\label{s:predgraph}
\subsection{Infinite Use of Finite Means}
\label{s:predgraph:infinite}
Chomsky was famously fond of quoting Humboldt's aphorism that language makes {\em infinite use of finite means} \cite{Chomsky1965Aspects}.
The {\em implication graph} allows us to estimate probabilities for an {\em unbounded} number of {\em propositions} $\pvariable$ based on finite parameters $\Psi$, by defining weights over {\em predicate patterns}, rather than relationships between {\em concrete entities}.
That is, we learn a general link between $\xjack$ {\em liking} $\xjill$ and $\xjack$ {\em dating} $\xjill$, and this can apply to $\cvariable_{jack1}$ or $\cvariable_{jack2}$ or $\cvariable_{jill1}$ or $\cvariable_{jill2}$, etc., and so make {\em infinite use} of {\em finite means}.

\subsection{Graph Operations}
\paragraph{Construction}
The {\em implication graph} is constructed from the set of all relevant {\em conjoined predicate implications} that we want to train weights for in our model:
\begin{equation}
    \mathcal{K} = \left\{\Psi(\hvariable, \qvariable)\right\}
\end{equation}

\paragraph{Backwards Links for a Predicate}
From this, we can recover the {\em backwards} set of all {\em predicate implication links} for a predicate $\qvariable$:
\begin{equation}
    B_\Psi(\qvariable) = \left\{\Psi(\hvariable, \qvariable') \in \mathcal{K}  \condsep \qvariable' = \qvariable \right\}
\end{equation}
It is also possible to define a {\em forwards} set but we avoid doing this and consider only $B_\Psi$ for simplicity.

\subsection{Abstraction and Backwards Substitution}
\paragraph{Abstraction}
For any {\em proposition} \pvariable, whose role set is $\rvariable$, we can {\em abstract} any subset of the roles in $\rvariable$ to reveal a predicate $\qvariable$.
For example, for the proposition:
\begin{equation} \pvariable = (\textsc{like}, \left\{\textsc{subj}: \constant{jack1}, \textsc{dobj}: \constant{jill1} \right\})\end{equation}
Abstracting $\left\{\textsc{subj}, \textsc{dobj}\right\}$ would leave:
\begin{equation}
    \qvariable = (\textsc{like}, \left\{\textsc{subj}: \xjack, \textsc{dobj}: \xjill \right\})
    \label{eq:}
\end{equation}
Though {\em abstracting} over all variables at once can be considered the {\em standard abstraction}, we can abstract partially
in $2^n - 1$ different ways, as $\pvariable$ is not included as an {\em abstraction} of itself, because it has no open roles.
We will write that $\qvariable \in \pvariable$ if $\qvariable$ is an abstraction of $\pvariable$.

\paragraph{Backwards Substitution}
Suppose that $\qvariable$ is an abstraction of $\pvariable$, i.e. $\qvariable \in \pvariable$.
And, suppose that $\Psi(\hvariable, \qvariable)$ is an implication link.
We can define:
\begin{equation}
    backfill(\pvariable, \Psi(\hvariable, \qvariable)) = \text{unique }\gvariable\text{ such that }\Psi(\hvariable, \qvariable)\text{ links $\gvariable$ to } \pvariable
\end{equation}
This function can be computed because we stored the {\em role mapping pair} $\left\{r, s\right\}$ for each $\qvariable_a \in \hvariable$ and $\qvariable$, for each $\Psi(\hvariable, \qvariable)$ in the implication graph.

\subsection{Proposition Factors and Contexts}
\paragraph{Proposition Factor}
Suppose we have a proposition $\pvariable$, which contains the abstraction $\qvariable$, which matches an implication link $\Psi(\hvariable, \qvariable)$.
We can call 
$backfill(\pvariable, \Psi(\hvariable, \qvariable))$ to obtain some $\gvariable$, an instance of $\hvariable$, obtained by following backwards an instance of the link $\Psi(\hvariable, \qvariable)$.
$\gvariable = \pvariable_1 \land ... \land \pvariable_n$ is a conjunction of propositions, and so has a {\em probability}, unlike $\hvariable$, which is a predicate. These objects are all bundled up in a {\em factor} defined as:
\begin{equation}
    factor(\pvariable, \Psi(\hvariable, \qvariable)) = (\pvariable, \Psi(\hvariable, \qvariable), \gvariable)
\end{equation}
The {\em factor} contains both the causally related proposition group $\gvariable$, and also the {\em implication link} $\Psi(\hvariable, \qvariable)$ used to link $\gvariable$ and $\pvariable$.

\paragraph{Proposition Factor Context}
For a given proposition $\pvariable$, its {\em factor context} is:
\begin{equation}
    \textsc{context}(\pvariable) =
        \bigcup_{\substack{\qvariable \in \pvariable}}\ 
            \bigcup_{\substack{\hvariable \in \backlinks (\qvariable)}}
            factor(\pvariable, \Psi(\hvariable, \qvariable))
\end{equation}
This is the set of all {\em factors} created from taking all {\em backwards implication links} from all {\em abstracted predicates} $\qvariable \in \pvariable$.
The factor context is the input to the learned {\em linear exponential} model used to score the \opor\ gates, $\Psi_\opor$.
\paragraph{Markov Assumption}
In terms of the {\em Markov assumption}, the node $\pvariable$ is independent of all its ancestors given its {\em factor context}.
That is, the {\em factor context} contains the set of all {\em direct causes} for $\pvariable$, according to the current {\em theory}.

\subsection{Inference-Time Proposition Graph Creation}
When we are interested in a query $\pvariable$, we have to construct the graph of relevant proprositions {\em on the fly} at inference time, because we cannot store all propositions.
Suppose we are interested in a certain target query $\pvariable$, which for simplicity for now assume has only ancestors, and no descendents in the graph.
We can determine $\textsc{context}(\pvariable)$, which will get all of the conjoined nodes $\gvariable$ that are {\em parents} of $\pvariable$.
Each $\gvariable_z = \pvariable_{z_1} \wedge ... \wedge \pvariable_{z_n}$ is a conjunction of $\pvariable_{z_i}$, and for each of these we can recursively call $\textsc{context}(\pvariable_{z_i})$, and so on, until we have created a {\em proposition graph} of all propositions {\em relevant to} $\pvariable$.
Because of the {\em Markov assumption}, any node not reached through this traversal is not relevant to $\pvariable$.
During the construction of this graph, we can do book-keeping to store, for each $\pvariable$ and $\gvariable$ discovered, the {\em forward links}, linking a node to its {\em children}, and {\em backward links}, linking a node to its {\em parents}, for each node of each type.
We use this all at {\em inference} time, described in Section \ref{sec:inference}.

\subsection{Feature Function}
\label{s:feature_function}
The feature function $\phi(\pvariable, \gvariable)$ characterizes the {\em implication link} between the conclusion $\pvariable$ and the assumption $\gvariable$:
\begin{equation}
    \phi(\pvariable = p, \gvariable = g) = (p, \Psi(\hvariable, \qvariable), g) 
\end{equation}
That is, the feature $\phi(\pvariable = p, \gvariable = g)$ is a {\em triple} indicating:
\begin{enumerate}
    \item The value $p \in \left\{0, 1\right\}$ that $\pvariable$ takes in $\phi(\pvariable = p, \gvariable = g)$.
    \item The implication link $\Psi(\hvariable, \qvariable)$ used to arrive at $\pvariable$ from $\gvariable$.
    \item The value $g \in \left\{0, 1\right\}$ that $\gvariable$ takes on in $\phi(\pvariable = p, \gvariable = g)$.
\end{enumerate}
We usually just write $\phi(\pvariable, \gvariable)$, and assume that the $\Psi(\hvariable, \qvariable)$ is implied.
It {\em is} possible for the same $\pvariable$ and $\gvariable$ to have more than one link, which would result in more than one feature.
The {\em feature vector} for the entire {\em factor context} is the union over each of the individual {\em proposition factors}.
\section{Inference}
\label{sec:inference}
\subsection{The Probability Query}
We are interested in the \textit{probability query}, which consists of two parts:
\begin{itemize}
    \item The {\em query variables}: a subset \( \pvariableset_Q \) of all variables in the network.
    \item The {\em evidence}: a subset \( \pvariableset_E \) of random variables in the network, {\em observed} to have the values \( \left\{p\right\}_E\).
\end{itemize}
The task is to compute the {\em posterior distribution}:
\begin{equation}
    P( \pvariableset_Q \mid \pvariableset_E = \left\{p\right\}_E)
\end{equation}

\subsection{Marginalization}
In the presence of unobserved variables \( \pvariableset_U \), not part of the query or evidence, marginalization is used to sum out these variables from the joint probability distribution. The marginalization process is represented by the following equation:
\begin{equation}
    P(\pvariableset_Q \mid \pvariableset_E) = \sum_{\pvariableset_U} P(\pvariableset_Q, \pvariableset_U \mid \pvariableset_E)
\end{equation}
In general, in a Bayesian Network, this process if $\Omega(2^N)$ to compute {\em exactly}, or even to {\em provably approximate} \cite{Cooper1990,Roth1996HardnessApproxReasoning}.

\subsection{Iterative Belief Propagation}
Inference can be performed in a graphical model using {\em belief propagation} \cite{koller2009probabilistic,neapolitan2003learning,bishop2006pattern}, if the graph contains even {\em undirected cycles}, which it often would, {\em exact} belief propagation is not tractable.
However, empirical results suggest that {\em loopy belief propagation}, which we will call {\em iterative belief propagation}, does converge well empirically, even though there are no theoretical guarantees \cite{murphy1999loopy, Smith2008}.
We discuss the complexity of this operation in detail in Section \ref{sec:complexity} and our results on convergence in Section \ref{sec:experiments}.

\subsection{Message Passing Calculations}
\paragraph{Notation}
We implement the variant of \cite{pearl1988probabilistic}'s {\em belief propagation} algorithm presented in \cite{neapolitan2003learning}.
In this formulation, we have $\pi$ {\em values} and $\lambda$ {\em values}, and $\pi$ {\em messages} and $\lambda$ {\em messages}.
For factor computations, we distinguished between {\em single propositions} $\pvariable$ and {\em proposition groups} $\gvariable$.
However, for the message passing calculations we adopt a unified notation, where both $\pvariable$ and $\gvariable$ nodes can be viewed as a unified node $\zvariable$ that can wrap either type, and the message passing calculations are agnostic to the type.
We use $\cvariable$ to canonically refer to a {\em child} of $\zvariable$ and $\avariable$ for a {\em parent} of $\zvariable$.
\paragraph{Computations}
The version we present here involves exponential cost $O(2^n)$ sums over either the parents or children of $z$.
In Section \ref{sec:complexity}, we discuss how the {\em independence} of $\Psi_\opor$ factors can, for some distributions like {\em Noisy Or}, allow the $O(2^n)$ update to be done in {\em linear} $O(n)$ time. 
\paragraph{Values}
$\pi(z)\in \mathbb{R}$, called the $\pi$ {\em value} for $\zvariable = z$, represents beliefs flowing {\em forward} in the network, from {\em causes} to {\em effects}, and is:
\begin{equation}
    \pi(z) = \sum_{a_1, \ldots, a_n} \left( P(z \mid a_1, \ldots, a_n) \prod_{a_i} \pi_\zvariable(a_i) \right).
    \label{eq:slow_pi_calc}
\end{equation}
$\lambda(z)\in \mathbb{R}$, called the $\lambda$ {\em value} for $\zvariable = z$, represents beliefs flowing {\em backward} in the network, from {\em effects} to {\em causes}, and is:
\begin{equation}
    \lambda(z) = \prod_{\cvariable} \lambda_{\cvariable}(z)
\end{equation}
These two values are normalized and combined to compute the {\em posterior} probability:
\begin{equation}P(z \condsep \left\{p\right\}_E) = \alpha\lambda(z)\pi(z) \end{equation}

\paragraph{Messages}
$\pi_\zvariable(a) \in \mathbb{R}$ is $\avariable$'s message to a {\em child} $\zvariable$:
\begin{equation}
    \pi_\zvariable(a) = \pi(a) \prod_{(\yvariable \in \zvariable) - \avariable} \lambda_\yvariable(z)
\end{equation}
$\lambda_\cvariable(z) \in \mathbb{R}$ is $\cvariable$'s message to a {\em parent} $\zvariable$, where the ${\bf b}_i$ are the other parents of $\cvariable$:
\begin{equation}
    \lambda_\cvariable(z) = \sum_{c} \left[ \sum_{b_1, b_2, \ldots, b_n} \left( P(c \mid z, b_1, b_2, \ldots, b_n) \prod_{b_i} \pi_\cvariable(b_i) \right) \lambda(c) \right]
    \label{eq_slow_lambda_calc}
\end{equation}

\section{Experiments}
\label{sec:experiments}
\subsection{Logical Structures}
\subsubsection{Method}
\paragraph{Synthetic Data}
We train the model with synthetic data.
Our goal is to show that the {\em QBBN} can {\em learn} the model, and to investigate {\em inference} using {\em iterative belief propagation}.
\paragraph{Example Universe}
We investigate the problem of of our running example in which there are two variables from a bipartite set, $\xjack$ and $\xjill$, and we are interested whether $date(\xjack, \xjill)$.
This is the problem discussed in  Section \ref{sec:running_example}, and the graphical model for our {\em theory} of this universe is depicted in Figure \ref{fig:alternating_network}.
For any $\xjack$, $lonely(\xjack)$ is true with probability $30\%$.
For any $\xjill$, $exciting(\xjill)$ is true with probability $60\%$.
For any $\xjack, \xjill$, $like(\xjack, \xjill)$ iff $lonely(\xjack) \lor exciting(\xjill)$.
For any $\xjill, \xjack$, $like(\xjill, \xjack)$ is {\em true} is true with probability $40\%$.
For any $\xjack, \xjill$, $date(\xjack, \xjill)$ iff $like(\xjack, \xjill)$ {\em and} $like(\xjill, \xjack)$.
\paragraph{Training}
We train on 4096 randomly generated synthetic examples.
We use a very basic {\em stochastic gradient descent} implementation, in which the learning rate is fixed, without averaging.
There is some error in this simplistic estimate but these experiments are primarily to check the behavior of {\em iterative belief propagation}.
\paragraph{Belief Propagation Convergence}
In each case we: 1) set some evidence (possibly nothing), 2) do $k$ rounds of {\em iterative belief propagation}, where the number of rounds is plotted on the {\em x-axis} in all graphs.
In all cases, iteration $0$ shows the prior probability, after which we either {\em set an observed variable} or {\em do nothing}.
If we set an observed variable, we then do {\em fan out} message passing from the observed variable, which involves doing {\em lambda} backward message passing up the graph from the changed node first, and then {\em pi} forward message passing back down the graph from the roots, with each fan out counting as one iteration.
In this case, we see how the graph changes over iterations.
If we did not set an observed variable, then we just do rounds of full {\em forward}-{\em backward} passes, to observe that the network does not change without new information.

\subsubsection{Results}
\label{sec:results}
\begin{figure}[htp]
    \centering
    \centering
    \includegraphics[width=\linewidth]{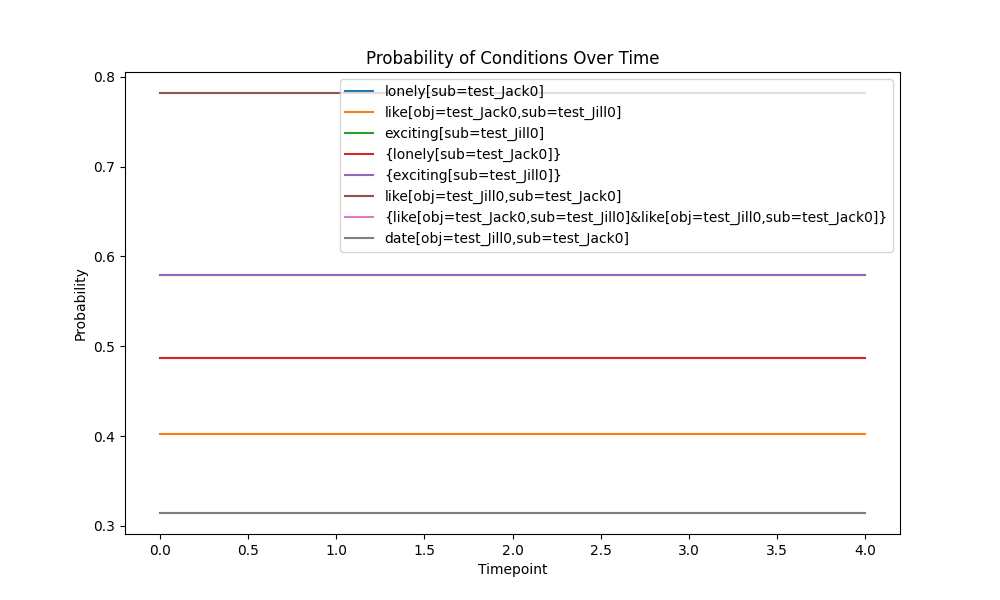}
    \caption{The {\em prior} state of the network, with no {\em observations}.}
    \label{fig:prior}
\end{figure}

\paragraph{No Evidence}
First, we investigate inference in the model for an example in which none of the variables are set.
Figure \ref{fig:prior} shows the baseline probabilities in the model, that match the by-hand calculations we can do to verify, with some noise due to the unsophisticated gradient descent.
$P(like(\xjack, \xjill))$ is a {\em noisy or} over $P(lonely(\xjack)) = 0.3$ and $exciting(\xjill) = 0.6$ so 
\[ P(like(\xjack, \xjill) = 1) = 1 - (1 - 0.3)(1 - 0.6) = 0.72 \]
In the network this is estimates as $0.78$, which we believe is due to the noise of the gradient descent.
$P(like(\xjack, \xjill))$ and $P(like(\xjill, \xjack))$ are {\em independent} (even in the underlying universe) so 
\[ P(like(\xjack, \xjill) \land like(\xjill, \xjack))  = P(like(\xjack, \xjill))\cdot P(like(\xjill, \xjack)) \]
This is $0.72 \cdot 0.4 =  0.29$, while the network the estimate is $0.31$.
We reiterate that we are primarily interested in the {\em message passing} in these experiments, and there are many well understood ways to improve the {\em SGD} estimate.

\begin{figure}[htp]
    \centering
    \centering
    \includegraphics[width=\linewidth]{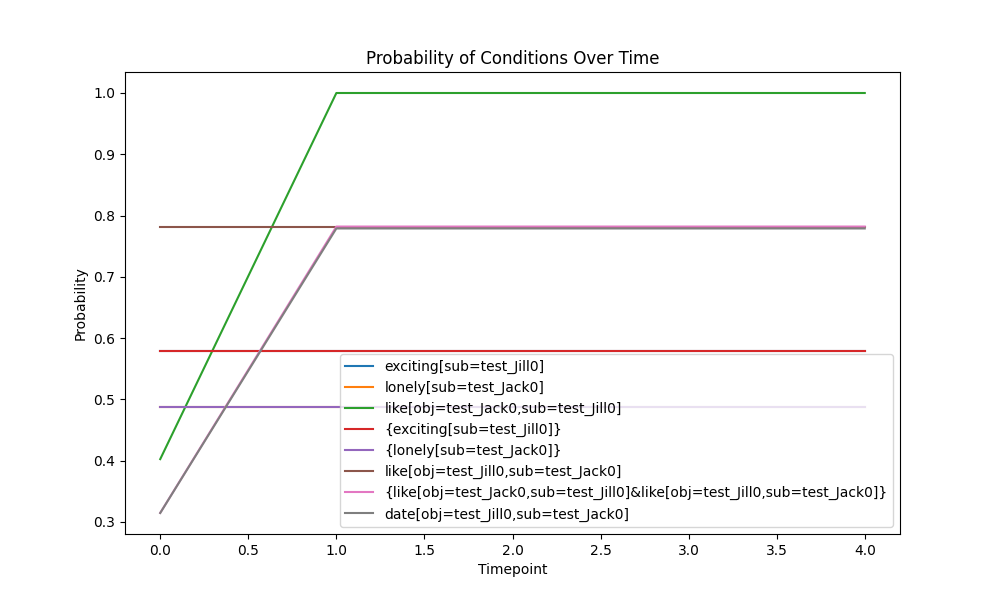}
    \caption{Assume that $like(\xjill, \xjack)$ is {\em true}. Forward inferences only.}
    \label{fig:jill_likes}
\end{figure}

\begin{figure}[htp]
    \centering
    \centering
    \includegraphics[width=\linewidth]{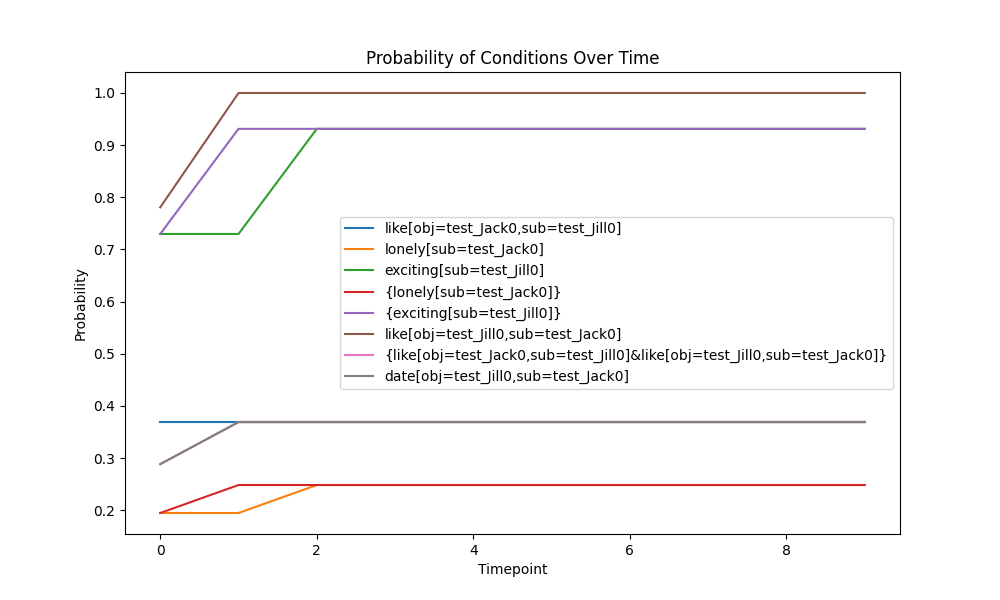}
    \caption{Assume that $like(\xjack, \xjill)$ is {\em true}. Forward and backwards inference.}
    \label{fig:jack_likes}
\end{figure}

\begin{figure}[htp]
    \centering
    \centering
    \includegraphics[width=\linewidth]{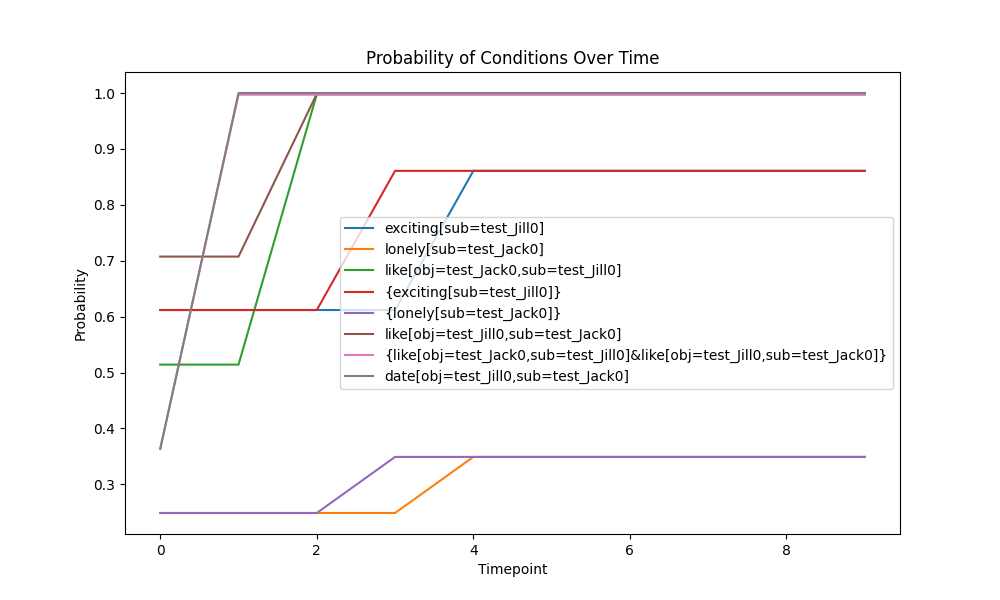}
    \caption{Assume that $date(\xjill, \xjack)$ is {\em true}. Backward inferences only.}
    \label{fig:they_date}
\end{figure}

\paragraph{Forward Only}
In Figure \ref{fig:jill_likes}, we assume that $like(\xjill, \xjack)$, which affects $P(date(\xjack, \xjill))$, but not $P(like(\xjack, \xjill))$, which is independent, and those so are its ancestors.
\paragraph{Forward and Backward}
In Figure \ref{fig:jack_likes}, we assume that $like(\xjack, \xjill)$, which affects both its parents and its children (which includes many variables), but not $like(\xjill, \xjack)$, which is independent.
\paragraph{Backward Only}
In Figure \ref{fig:they_date}, we assume that $date(\xjack, \xjill)$, which backwards infers through the $\Psi_\opand$ gate, to $like(\xjack, \xjill)$ and $like(\xjill, \xjack)$. The inference of $like(\xjack, \xjill)$ implies updated beliefs about its ancestors $lonely(\xjack)$ and $exciting(\xjill)$ as well.

\subsection{Message Propagation}
\paragraph{Method}
We have just seen that the {\em QBBN} with {\em iterative belief propagation} can infer over logical structures.
We now ask about the propagation of beliefs over {\em distance} in the graph.
To do this, we consider only a single variable $\xjack$, and a series of unary predicates $\alpha_0, ..., \alpha_N$, where we use $N = 10$.
Now $\alpha_0(\xjack)$ is determined by a $50\%$ cointoss.
Then, for $i \geq 1$, we deterministically set $\alpha_{i}(\xjack) = \alpha_{i-1}(\xjack)$.
That is, while $\alpha_0(\xjack)$ is random, $\alpha_i(\xjack)$ for $i \geq 1$ can be determined with certainty if we know the value of $\alpha_{i-1}(\xjack)$ or $\alpha_{i+1}(\xjack)$.
We examine how the beliefs change if we observe either $\alpha_0(\xjack)$ or $\alpha_N(\xjack)$ are observed to be true.

\begin{figure}[htp]
    \centering
    \centering
    \includegraphics[width=\linewidth]{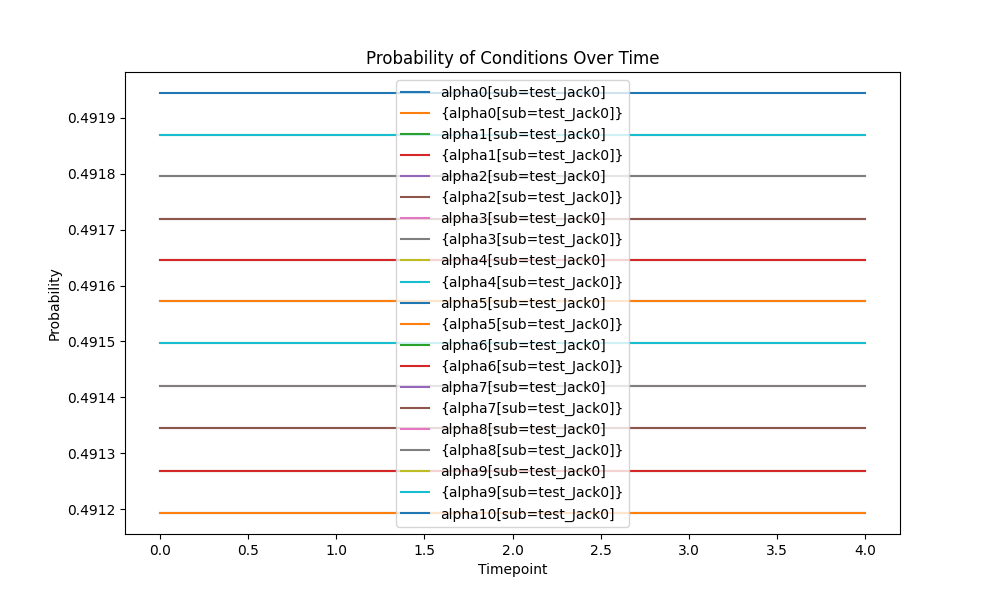}
    \caption{This shows the {\em prior} state of the $\alpha_i(\xjack)$ network. Before knowing anything at all, we expect $P(\alpha_i(\xjack)) = 0.5$ for all $i$.}
    \label{fig:long_chain_prior}
\end{figure}

\begin{figure}[htp]
    \centering
    \centering
    \includegraphics[width=\linewidth]{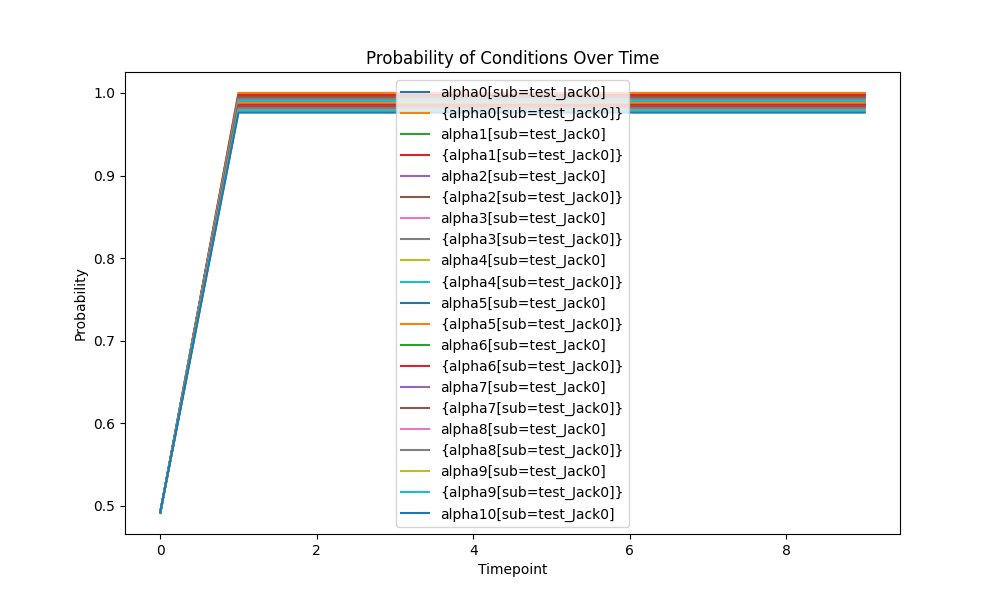}
    \caption{After observing $\alpha_0(\xjack) = 1$, the beliefs propagate {\em forward} in one pass.}
    \label{fig:long_chain_set_0_1}
\end{figure}

\begin{figure}[htp]
    \centering
    \centering
    \includegraphics[width=\linewidth]{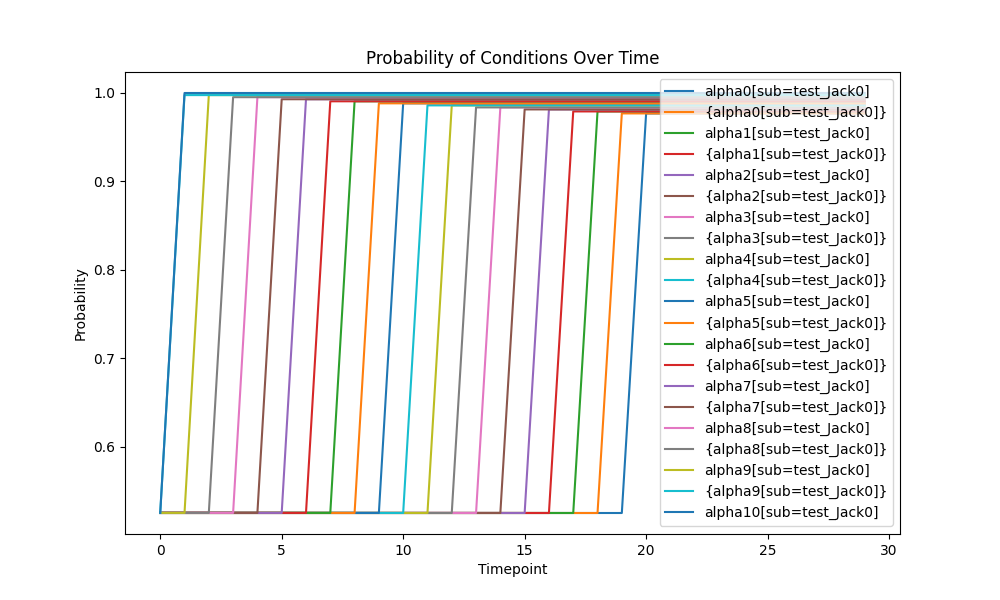}
    \caption{After observing $\alpha_N(\xjack) = 1$, the beliefs propagate {\em backwrd} at a rate of one new node changing per iteration, and note that there are intermediate {\em conjunction} nodes, adding a constant factor to the convergence time.}
    \label{fig:long_chain_set_n_1}
\end{figure}

\paragraph{Results}
Figure \ref{fig:long_chain_prior} shows the {\em prior} probability of each node in the network, which is around $50\%$, with some noise, as discussed above.
Figure \ref{fig:long_chain_set_0_1} shows what happens when we observe $\alpha_0(\xjack) = 1$, the information propagates {\em forward} through the network in {\em one} iteration {\em total}.
Figure \ref{fig:long_chain_set_n_1} shows what happens when we observe $\alpha_N(\xjack) = 1$, the information propagates {\em backward} through the network, at a rate of {\em one} new node chaning per iteration, taking {\em twenty} iterations in total..
Note that, for each variable $\alpha_i(\xjack)$, $i \geq 1$, there is also an intermediate ``conjoined'' node with only one element $\left\{\alpha_{i-1}(\xjack)\right\}_\land$.
We believe more efficient ways of managing belief propagation are possible than just doing repeated fan outs, but we leave this for future work.

\section{Complexity of Inference}
\label{sec:complexity}
\subsection{Provably Exact Inference}
Inference in a general Bayesian Network is $\Omega(2^N)$ for $N$ variables, and is even $\Omega(2^N)$ to {\em provably} approximate \cite{Cooper1990, Roth1996HardnessApproxReasoning}.
\subsection{Empirically Successful Iterative Belief Propagation}
The {\em iterative belief propagation} ({\em loopy belief propagation} in the literature) is {\em not} guaranteed to converge \cite{neapolitan2003learning, koller2009probabilistic}, but {\em has been found} to converge in practice in a range of studies \cite{Smith2008, murphy1999loopy, Gormley2015}.
And, we have found it to converge in our experiments.
The primary cost of {\em inference} in this case is the computation of the {\em messages and values} of the $\pi$ and $\lambda$ tables (see Section \ref{sec:inference})
Using a perhaps {\em naive} implementation, in which the marginalization is exact (see \ref{eq:slow_pi_calc} and \ref{eq_slow_lambda_calc}), runs in time $O(2^n)$ in $n$ the number of {\em inputs} to the {\em factor}.
Then, a single pass of belief propagation visits each of the $N$ nodes once, taking total time $O(N2^n)$, and empirically $k$ rounds are needed to converge.
However, it may be possible to make both $\Psi_\opand$ and $\Psi_\opor$ gates faster, as we now discuss.
\subsection{Faster Disjunction}
\label{sec:noisy_or}
\paragraph{Overview}
The $\Psi_\opor$ factor is {\em learned} when we want to do statistical inference, and the factor
\[\Psi_\opor(\pvariable \condsep \gvariable_1, ..., \gvariable_n)\]
has one input $n$ per {\em modeled cause} $\gvariable_i$ of $\pvariable$.
Conceptually, outcomes have an unbounded number of potential causes, and we would ideally not need to restrict $n$ solely because of message passing complexity.
\paragraph{Importance Sampling}
One option is to {\em learn} an unbounded number $n$ of weights, but only {\em consider at inference} a {\em subset} of the inputs that are most {\em relevant}.
That is, in the {\em linear exponential} model \ref{e_linear_exponential}, we can detect which of $m < n$ linear contributions will have the biggest effect, and only marginalize over those, costing $O(2^m) < O(2^n)$.
This strategy would be a variant of {\em importance sampling} \cite{wilkinson2005grammar}.
\paragraph{Linear Time Disjunction}
\cite{neapolitan2003learning} lists at least one disjunction model, the {\em Noisy Or} model, whose message passing calculations are $O(n)$, instead of $O(2^n)$ in $n$ the number of inputs to the factor.
We leave it to future work to determine whether this model, or another model with similar scaling properties, can be useful in practice.

\subsection{Faster Conjunction}
The complexity of {\em message updates} in a {\em conjunction} $\Psi_\opand$ gate is $O(2^n)$ in the number of inputs $n$.
However, an \opand\ gate can be arranged into a {\em binary tree} of \opand\ gates each of size $2$, with the tree height $\log_2(n)$, in which case there would be only $O(n)$ work in total to evaluate the $n$ inputs.
However, this would increase the amount of message passing, so we leave it to future work to evaluate whether this is beneficial.
\section{Compared to Other Logical Models}
\paragraph{AlphaGeometry}
\cite{Trinh2024} present a model that is trained to solve problems from the {\em geometry olympiad}.
We observe that such questions are purely mathematical, and so can be solved by ordinary {\em deterministic theorem provers}.
So, the use of {\em LLM}'s would seem to us a potential {\em efficiency improvement} in the field of automatic theorem-proving, which is definitely an interesting direction to consider.
Our work focuses instead on the {\em general relationship} between {\em logical} and {\em statistical reasoning}.
Our analysis of {\em simple} versus {\em complex} kinds of inferences \cite{Coppola2024}, shows that there is a difference between {\em everyday} reasoning, and the kinds of {\em complex mathematical reasoning} that is necessary for mathematics: specifically, in {\em complex proofs}, the {\em assumptions change}, and this requires some kind of {\em book-keeping} or {\em resoning by cases}.
This shows why we may {\em not} want to do full theorem-proving for typical {\em information retrieval}, but instead focus on only those inferences that are {\em fast}.
Also, our work is meant to apply to {\em all} language, and to analyze the logical and probabilistic nature of all language.
Also, our analysis of the {\em QBBN} in terms of a {\em complete and consistent} calculus on the basis of \cite{Prawitz1965} allows us to understand what logical rules are implemented now, versus what {\em can be} implemented, and thus provides a clear, principled program for further research.

\paragraph{Self-Discover}
\cite{Zhou2024} uses a technique called {\em self-discover} to learn what the authors say is {\em cause-and-effect reasoning}.
This work seems to rely on {\em modules} of computation, that seem to be complex but not listed.
Also, it is not clear how this method would be extended to ensure {\em logical} or {\em probabilistic} {\em consistency} in a graph containing one variable per {\em possible} proposition, which is an enormous and indeed unbounded number.
There are no {\em complex modules} in our work, all of the equations are given here.
Also, our use of the well-studied system of {\em Bayesian Networks} \cite{pearl1988probabilistic}, gives us a framework for understanding {\em consistency} that is non-trivial and contains many useful results and theorems.

\section{Implementation}
An implementation of the {\em QBBN} called {\em BAYES STAR} can be found at \cite{CoppolaBayesStar2024}.
This implementation is written in the {\em Rust} programming language, with {\em REDIS} for storage, and includes the code for training and doing inference in each of the examples discussed here.
\section{Future Work}
\paragraph{Learning from Unlabeled Text}
We have said that the {\em QBBN} can encode knowledge, and do so {\em without hallucinating}, which compares favorably with the {\em LLM} \cite{Bahdanau2014NeuralMT, Vaswani2017, radford2018improving}.
However, the difficulty compared to the {\em LLM} is that the {\em QBBN} {\em cannot} be learned in the same {\em direct} {\em $n$-gram language model} way as the {\em LLM}, but instead must refer to {\em logical forms} which are {\em not observed} but viewed as {\em latent} and must be learned through {\em expectation maximization} \cite{dempster1977maximum}.
\paragraph{Belief Propagation}
We have used {\em loopy belief propagation}, calling it {\em iterative belief propagation}, which is not guaranteed to converge but has been studied somewhat extensively \cite{murphy1999loopy, Smith2008, Gormley2015}, and our experiments also find convergence.
However, convergence for larger graphs should be studied, as well as strategies to speed up belief propagation.
\paragraph{Logical Language Features}
We have shown enough about the underlying logical language of the {\em QBBN} to encode {\em basic} first-order sentences.
But, there remain the topics of {\em compositional semantics} \cite{montague1970universal}, which shows how the meanings of {\em larger parts} are made from {\em smaller parts}, and {\em intensional} semantics \cite{montague_proper_treatment}, which shows how {\em the concept} behind a sentence can itself be an argument.
\bibliographystyle{apalike}
\bibliography{bibtex}
\end{document}